\documentclass[11pt,a4paper]{article}
\usepackage[hyperref]{acl2021}
\usepackage{times}
\usepackage{latexsym}
\usepackage{comment}
\usepackage{graphicx}

\usepackage{caption}
\usepackage{subcaption}
\usepackage{float}

\usepackage{microtype}
\usepackage{colortbl}
\usepackage{booktabs}

\aclfinalcopy % Uncomment this line for the final submission
\def\aclpaperid{32} %  Enter the acl Paper ID here

\newcommand\fb{\texttt{FB}}
\newcommand\pin{\texttt{Pin}}

\title{
Memes in the Wild: Assessing the Generalizability \\ of the Hateful Memes Challenge Dataset}

\author{{\small Hannah Rose Kirk$^{1\dagger\ddagger}$, Yennie Jun$^{1\dagger}$, Paulius Rauba$^{1\dagger}$, Gal Wachtel$^{1\dagger}$, Ruining Li$^{2\dagger}$},\\
 {\small {\bfseries Xingjian Bai$^{2\dagger}$, Noah Broestl$^{3\dagger}$, Martin Doff-Sotta$^{4\dagger}$, Aleksandar Shtedritski$^{4\dagger}$, Yuki M. Asano$^{4\dagger}$}}\\
  { \small $^1$Oxford Internet Institute, $^2$Dept. of Computer Science, $^3$Oxford Uehiro Centre for Practical Ethics} \\ 
  {\small  $^4$Dept. of Engineering Science, $^\dagger$ Oxford Artificial Intelligence Society} \\
  {\small $^\ddagger$hannah.kirk@oii.ox.ac.uk}\\
}

\date{}

\begin{document}
\maketitle
\begin{abstract}
% The generalizability of Facebook's Hateful Meme Challenge has consequences for detecting hateful memes `in the wild'. 

Hateful memes pose a unique challenge for current machine learning systems because their message is derived from both text- and visual-modalities.
To this effect, Facebook released the Hateful Memes Challenge, a dataset of memes with pre-extracted text captions, but it is unclear whether these synthetic examples generalize to `memes in the wild'.
In this paper, we collect hateful and non-hateful memes from Pinterest to evaluate out-of-sample performance on models pre-trained on the Facebook dataset.
We find that memes in the wild differ in two key aspects:
1) Captions must be extracted via OCR, injecting noise and diminishing performance of multimodal models, and
2) Memes are more diverse than `traditional memes', including screenshots of conversations or text on a plain background.
This paper thus serves as a reality check for the current benchmark of hateful meme detection and its applicability for detecting real world hate. 
\end{abstract}

\section{Introduction}
Hate speech is becoming increasingly difficult to monitor due to an increase in volume and diversification of type \citep{macavaney2019hate}. 
% To standardize performance
To facilitate the development of multimodal hate detection algorithms, Facebook introduced the Hateful Memes Challenge, a dataset synthetically constructed by pairing text and images \citep{Kiela2020TheHM}.
Crucially, a meme's hatefulness is determined by the combined meaning of image and text.
The question of likeness between synthetically created content and naturally occurring memes is both an ethical and technical one:
Any features of this benchmark dataset which are not representative of reality will result in models potentially overfitting to `clean' memes and generalizing poorly to memes in the wild.
Thus, we ask the question: How well do Facebook's synthetic examples (\fb) represent memes found in the real world? 
We use Pinterest memes (\pin) as our example of memes in the wild and explore differences across three aspects:
\begin{enumerate}
    \item \textbf{OCR.} While \fb~memes have their text pre-extracted, memes in the wild do not. Therefore, we test the performance of several Optical Character Recognition (OCR) algorithms on \pin~and \fb~memes.
    
    \item \textbf{Text content}. To compare text modality content, we examine the most frequent n-grams and train a classifier to predict a meme's dataset membership based on its text.
    
    \item \textbf{Image content and style}. To compare image modality, we evaluate meme types (traditional memes, text, screenshots) and attributes contained within memes (number of faces and estimated demographic characteristics).
\end{enumerate}
After characterizing these differences, we evaluate a number of unimodal and multimodal hate classifiers pre-trained on \fb~memes to assess how well they generalize to memes in the wild. 

\section{Background}
The majority of hate speech research focuses on text, mostly from Twitter \citep{waseem2016, davidson2017, founta2018large, zampieri2019}. Text-based studies face challenges such as distinguishing hate speech from offensive speech \citep{davidson2017} and counter speech \citep{mathew2018}, as well as avoiding racial bias \citep{sap2019risk}. Some studies focus on \textit{multimodal} forms of hate, such as sexist advertisements \citep{gasparini2018multimodal}, YouTube videos \citep{poria2016fusing}, and memes \citep{suryawanshi2020multimodal,  zhou2020multimodal, das2020detecting}. 

While the Hateful Memes Challenge \citep{Kiela2020TheHM} encouraged innovative research on multimodal hate, many of the solutions may not generalize to detecting hateful memes at large. For example, the winning team~\citet{zhong2020classification} exploits a simple statistical bias resulting from the dataset generation process.
While the original dataset has since been re-annotated with fine-grained labels regarding the target and type of hate \citep{nie_2021}, this paper focuses on the binary distinction of hate and non-hate. 

\section{Methods}
\subsection{Pinterest Data Collection Process} 
Pinterest is a social media site which groups images into collections based on similar themes. The search function returns images based on user-defined descriptions and tags. Therefore, we collect memes from Pinterest\footnote{We use an open-sourced Pinterest scraper, available at \url{https://github.com/iamatulsingh/pinterest-image-scrap}.} using keyword search terms as noisy labels for whether the returned images are likely hateful or non-hateful (see Appendix \ref{sec:appendix_data}). For hate, we sample based on two heuristics: synonyms of hatefulness or specific hate directed towards protected groups (e.g., `offensive memes', `sexist memes') and slurs associated with these types of hate (e.g., `sl*t memes', `wh*ore memes'). For non-hate, we again draw on two heuristics: positive sentiment words (e.g., `funny', `wholesome', `cute') and memes relating to entities excluded from the definition of hate speech because they are not a protected category (e.g., `food', `maths'). Memes are collected between March 13 and April 1, 2021. We drop duplicate memes, leaving 2,840 images, of which 37\% belong to the hateful category.

\subsection{Extracting Text- and Image-Modalities (OCR)} 
We evaluate the following OCR algorithms on the \pin~and \fb~datasets: Tesseract \citep{smith2007overview}, EasyOCR \citep{easy} and East \citep{east}.
Previous research has shown the importance of prefiltering images before applying OCR algorithms \citep{Bieniecki2007ImagePF}. Therefore, we consider two prefiltering methods fine-tuned to the specific characteristics of each dataset (see Appendix \ref{sec:appendix_OCR}). 
\subsection{Unimodal Text Differences}
After OCR text extraction, we retain words with a probability of correct identification $\ge 0.5$, and remove stopwords. A text-based classification task using a unigram Naïve-Bayes model is employed to discriminate between hateful and non-hateful memes of both \pin~and \fb~datasets. 

\subsection{Unimodal Image Differences}
To investigate the distribution of \textit{types} of memes, we train a linear classifier on image features from the penultimate layer of CLIP (see Appendix \ref{app:train_classifier}) \citep{radford2021learning}. From the 100 manually examined \pin~memes, we find three broad categories: 1) traditional memes; 2) memes consisting of just text; and 3) screenshots. Examples of each are shown in Appendix \ref{app:train_classifier}.  Further, to detect (potentially several) human faces contained within memes and their relationship with hatefulness, we use a pre-trained FaceNet model \citep{facenet} to locate faces and apply a pre-trained DEX model \citep{dex} to estimate their ages, genders, races. We compare the distributions of these features between the hateful/non-hateful samples.

We note that these models are controversial and may suffer from algorithmic bias due to differential accuracy rates for detecting various subgroups. \citet{Alvi2018TurningAB} show DEX contains erroneous age information, and \citet{Terhorst2021ACS} show that FaceNet has lower recognition rates for female faces compared to male faces. These are larger issues discussed within the computer vision community \citep{buolamwini2018gender}.

\subsection{Comparison Across Baseline Models}
To examine the consequences of differences between the \fb~and \pin~datasets, we conduct a preliminary classification of memes into hate and non-hate using benchmark models.
First, we take a subsample of the \pin~dataset to match Facebook's \texttt{dev} dataset, which contains 540 memes, of which 37\% are hateful. We compare performance across three samples: (1) \fb~memes with `ground truth' text and labels; (2) \fb~memes with Tesseract OCR text and ground truth labels; and (3) \pin~memes with Tesseract OCR text and noisy labels. Next, we select several baseline models pretrained on \fb~memes\footnote{These are available for download at \url{https://github.com/facebookresearch/mmf/tree/master/projects/hateful_memes}.}, provided in the original Hateful Memes challenge \citep{Kiela2020TheHM}. Of the 11 pretrained baseline models, we evaluate the performance of five that do not require further preprocessing: Concat Bert, Late Fusion, MMBT-Grid, Unimodal Image, and Unimodal Text. We note that these models are not fine-tuned on \pin~memes but simply evaluate their transfer performance. 
Finally, we make zero-shot predictions using CLIP \citep{radford2021learning}, and evaluate a linear model of visual features trained on the \fb~dataset (see Appendix \ref{app:train_clip}).

\section{Results}
\subsection{OCR Performance}
% \note{Martin}
Each of the three OCR engines is paired with one of the two prefiltering methods tuned specifically to each dataset, forming a total of six pairs for evaluation. For both datasets, the methods are tested on 100 random images with manually annotated text. For each method, we compute the average cosine similarity of the joint TF-IDF vectors between the labelled and cleaned\footnote{The cleaned text is obtained with lower case conversion and punctuation removal.} predicted text, shown in Tab.~\ref{tab:ocr}. Tesseract with \fb~tuning performs best on the \fb~dataset, while Easy with \pin~tuning performs best on the \pin~dataset. We evaluate transferability by comparing how a given pair performs on both datasets. \textbf{OCR transferability is generally low}, but greater from the \fb~dataset to the \pin~dataset, despite the latter being more general than the former. This may be explained by the fact that the dominant form of \pin~memes (i.e. text on a uniform background outside of the image) is not present in the \fb~dataset, so any method specifically optimized for \pin~memes would perform poorly on \fb~memes.
\begin{table}[h]
\setlength{\tabcolsep}{2pt}
\footnotesize
\centering
\caption{Cosine similarity between predicted text and labelled text for various OCR engines and prefiltering pairs. 
Best result per dataset is bolded.}
\label{tab:ocr}
\begin{tabular}{l ccc}
\toprule
                            & \textbf{\fb} & \textbf{\pin}                    & $|\Delta |$\\ 
\midrule
Tesseract, FB tuning        & \textbf{0.70}   & 0.36           & 0.34   \\ 
Tesseract, Pin tuning & 0.22             & 0.58           & 0.26   \\ 
Easy, FB tuning             & 0.53             & 0.30           & 0.23    \\ 
Easy, Pin tuning      & 0.32             & \textbf{0.67} & 0.35   \\ 
East, FB tuning             & 0.36             & 0.17           & 0.19   \\ 
East, Pin tuning      & 0.05             & 0.32           & 0.27   \\
\bottomrule
\end{tabular}
\end{table}

\subsection{Unimodal Text Differences}
We compare unigrams and bigrams across datasets after removing stop words, numbers, and URLs. The bigrams are topically different (refer to Appendix \ref{app:bigrams}).
A unigram token-based Naïve-Bayes classifier % (using add-one smoothing)
is trained on both datasets separately to distinguish between hateful and non-hateful classes. The model achieves an accuracy score of 60.7\% on \fb~memes and 68.2\% on \pin~memes (random guessing is 50\%), indicating mildly different text distributions between hate and non-hate. In order to understand the differences between the type of language used in the two datasets, a classifier is trained to discriminate between \fb~and \pin~memes (regardless of whether they are hateful) based on the extracted tokens. The accuracy is 77.4\% on a balanced test set. The high classification performance might be explained by the OCR-generated junk text in the \pin~memes which can be observed in a t-SNE plot (see Appendix \ref{app:tsne}).

\subsection{Unimodal Image Differences}
While the \fb~dataset contains only ``traditional memes''\footnote{The misclassifications into other types reflect the accuracy of our classifier.}, we find this definition of `a meme' to be too narrow: \textbf{the \pin~memes are more diverse}, containing 15\% memes with only text and 7\% memes which are screenshots (see Tab.~\ref{tab:meme_types}).
\begin{table}[h]
\centering
\footnotesize
\caption{Percentage of each meme type in \pin~and \fb~datasets, extracted by CLIP.}
\label{tab:meme_types}
\begin{tabular}{lccc} 
\toprule
  Meme Type                  & \textbf{\fb} & \textbf{\pin} & $|\Delta |$ \\
\midrule 
Traditional meme & 95.6\%                 & 77.3\%    & 18.3\%            \\
Text    & 1.4\%                 & 15.3\%     &13.9\%          \\
Screenshot & 3.0\%                 & 7.4\%         &4.4\%      \\
\bottomrule
\end{tabular}
\end{table}

\noindent Tab.~\ref{tab:facial_analysis} shows the facial recognition results. 
\textbf{We find that \pin~memes contain fewer faces than \fb~memes}, while other demographic factors broadly match.
% Hateful memes tend to have more faces than non-hateful memes. 
The DEX model identifies similar age distributions by hate and non-hate and by dataset, with an average of 30 and a gender distribution heavily skewed towards male faces (see Appendix \ref{app:face_recognition} for additional demographics).

\begin{table}[!htbp]
\caption{Facial detection and demographic (gender, age) distributions from pre-trained FaceNet and DEX.}
\label{tab:facial_analysis}
\centering
\footnotesize
% \resizebox{\columnwidth}{!}{%
\setlength{\tabcolsep}{2pt}
\begin{tabular}{l cccc}
\toprule
& \multicolumn{2}{c}{\textbf{\fb}} & \multicolumn{2}{c}{\textbf{\pin}}  \\ 
\textit{metric} & Hate   & Non-Hate   & Hate  & Non-Hate \\ 
\midrule 
Images w/ Faces   & 72.8\% & 71.9\% & 52.0\% & 38.8\% \\ 
Gender (M:F)      & 84:16 & 84:16 & 82:18 & 88:12\\ 
Age               & 30.7$_{\pm5.7}$ &31.2$_{\pm6.3}$ &29.4$_{\pm5.5}$ &29.9$_{\pm5.4}$ \\ 
\bottomrule
\end{tabular}
% }
\end{table}

\subsection{Performance of Baseline Models}
How well do hate detection pipelines generalize? Tab.~\ref{tab:model f1 scores} shows the F1 scores for the predictions of hate made by each model on the three samples: (1) \fb~with ground-truth caption, (2) \fb~with OCR, (3) \pin~with OCR. 

\begin{table}[h]
\centering
\footnotesize
% \setlength{\extrarowheight}{0pt}
% \addtolength{\extrarowheight}{\aboverulesep}
% \addtolength{\extrarowheight}{\belowrulesep}
\setlength{\aboverulesep}{0pt}
\setlength{\belowrulesep}{0pt}
\setlength{\tabcolsep}{4pt}
\caption{F1 scores for pretrained baseline models on three datasets. Best result per dataset is bolded.}
\label{tab:model f1 scores}
\begin{tabular}{lccc} 
\toprule
& \multicolumn{2}{c}{\textbf{\fb}} & \textbf{\pin} \\
Text from:             & Ground-truth        & OCR      & OCR          \\ 
\midrule
\multicolumn{4}{c}{{\cellcolor[rgb]{0.753,0.753,0.753}}\textbf{Multimodal Models}}    \\
Concat BERT   & 0.321          & 0.278          & 0.184                     \\
Late Fusion    & 0.499 & 0.471 & 0.377          \\
MMBT-Grid    & 0.396          & 0.328          & 0.351                     \\
\multicolumn{4}{c}{{\cellcolor[rgb]{0.753,0.753,0.753}}\textbf{Unimodal Models}}      \\
Text BERT  & 0.408          & 0.320          & 0.327                     \\
Image-Grid$^{*}$ & 0.226          & 0.226          & 0.351                     \\
\multicolumn{4}{c}{{\cellcolor[rgb]{0.753,0.753,0.753}}\textbf{CLIP Models}} \\
CLIP\textsubscript{Zero-Shot}$^{*}$ & 0.509 & 0.509 & 0.543 \\
CLIP\textsubscript{Linear Probe}$^{*}$ & \textbf{0.556}& \textbf{0.556} & \textbf{0.569} \\
\hline
\multicolumn{4}{p{\columnwidth}}{$^{*}$ these models do not use any text inputs so F1 scores repeated for ground truth and OCR columns.} \\
\bottomrule
\end{tabular} 
\end{table}

\textbf{Surprisingly, we find that the CLIP\textsubscript{Linear Probe} generalizes very well}, performing best for all three samples, with superior performance on \pin~memes as compared to \fb~memes. Because CLIP has been pre-trained on around 400M image-text pairs from the Internet, its learned features generalize better to the \pin~dataset, even though it was fine-tuned on the \fb~dataset.
%than to the synthetic Facebook memes. 
Of the multimodal models, Late Fusion performs the best on all three samples. When comparing the performance of Late Fusion on the \fb~and \pin~OCR samples, \textbf{we find a significant drop in model performance of 12 percentage points}. The unimodal text model performs significantly better on \fb~with the ground truth annotations as compared to either sample with OCR extracted text. This may be explained by the `clean' captions which do not generalize to real-world meme instances without pre-extracted text.

\section{Discussion}
The key difference in text modalities derives from the efficacy of the OCR extraction, where messier captions % versus clean pre-extracted text
result in performance losses in Text BERT classification. This forms a critique of the way in which the Hateful Memes Challenge is constructed, in which researchers are incentivized to rely on the pre-extracted text rather than using OCR; thus, the reported performance overestimates success in the real world. 
Further, the Challenge defines a meme as `a traditional meme' but we question whether this definition is too narrow to encompass the diversity of real memes found in the wild, such as screenshots of text conversations. % 

%The types of Pinterest memes are more diverse and heterogeneous. Specifically, there are more non-standard meme types, such as screenshots of text conversations. 
% While we initially hypothesized that models trained solely on Facebook memes would not generalize to the diversity of Pinterest meme types, the results in Tab.~\ref{tab:model f1 scores} show that Image-Grid has comparable accuracy across both Facebook and Pinterest memes. 

When comparing the performance of unimodal and multimodal models, we find multimodal models have superior classification capabilities which may be because the combination of multiple modes create meaning beyond the text and image alone \citep{Kruk2019IntegratingTA}. For all three multimodal models (Concat BERT, Late Fusion, and MMBT-Grid), the  score for \fb~memes with ground truth captions is higher than that of \fb~memes with OCR extracted text, which in turn is higher than that of \pin~memes. 
Finally, we note that CLIP's performance, for zero-shot and linear probing, surpasses the other models and is stable across both datasets.

\paragraph{Limitations}
Despite presenting a preliminary investigation of the generalizability of the \fb~dataset to memes in the wild, this paper has several limitations. Firstly, the errors introduced by OCR text extraction resulted in `messy' captions for \pin~memes. This may explain why \pin~memes could be distinguished from \fb~memes by a Naïve-Bayes classifier using text alone. However, these errors demonstrate our key conclusion that the pre-extracted captions of \fb~memes are not representative of the appropriate pipelines which are required for real world hateful meme detection.

Secondly, our \pin~dataset relies on noisy labels of hate/non-hate based on keyword searches, but this chosen heuristic may not catch subtler forms of hate. Further, user-defined labels introduce normative value judgements of whether something is `offensive' versus `funny', and such judgements may differ from how Facebook's community standards define hate \citep{fb_community_standards}. In future work, we aim to annotate the \pin~dataset with multiple manual annotators for greater comparability to the \fb~dataset. These ground-truth annotations will allow us to pre-train models on \pin~memes and also assess transferability to \fb~memes.

\paragraph{Conclusion}
We conduct a reality check of the Hateful Memes Challenge. Our results indicate that there are differences between the synthetic Facebook memes and `in-the-wild' Pinterest memes, both with regards to text and image modalities. Training and testing unimodal text models on Facebook's pre-extracted captions discounts the potential errors introduced by OCR extraction, which is required for real world hateful meme detection. We hope to repeat this work once we have annotations for the Pinterest dataset and to expand the analysis from comparing between the binary categories of hate versus non-hate to include a comparison across different types and targets of hate.

\bibliographystyle{acl_natbib}
\bibliography{references}

\begin{thebibliography}{29}
\expandafter\ifx\csname natexlab\endcsname\relax\def\natexlab#1{#1}\fi

\bibitem[{Alvi et~al.(2018)Alvi, Zisserman, and
  Nell{\aa}ker}]{Alvi2018TurningAB}
M.~Alvi, Andrew Zisserman, and C.~Nell{\aa}ker. 2018.
\newblock Turning a blind eye: Explicit removal of biases and variation from
  deep neural network embeddings.
\newblock In \emph{ECCV Workshops}.

\bibitem[{Bieniecki et~al.(2007)Bieniecki, Grabowski, and
  Rozenberg}]{Bieniecki2007ImagePF}
Wojciech Bieniecki, Szymon Grabowski, and Wojciech Rozenberg. 2007.
\newblock Image preprocessing for improving ocr accuracy.
\newblock In \emph{2007 international conference on perspective technologies
  and methods in MEMS design}, pages 75--80. IEEE.

\bibitem[{Buolamwini and Gebru(2018)}]{buolamwini2018gender}
Joy Buolamwini and Timnit Gebru. 2018.
\newblock Gender shades: Intersectional accuracy disparities in commercial
  gender classification.
\newblock In \emph{Conference on fairness, accountability and transparency},
  pages 77--91. PMLR.

\bibitem[{Das et~al.(2020)Das, Wahi, and Li}]{das2020detecting}
Abhishek Das, Japsimar~Singh Wahi, and Siyao Li. 2020.
\newblock Detecting hate speech in multi-modal memes.
\newblock \emph{arXiv preprint arXiv:2012.14891}.

\bibitem[{Davidson et~al.(2017)Davidson, Warmsley, Macy, and
  Weber}]{davidson2017}
Thomas Davidson, Dana Warmsley, Michael Macy, and Ingmar Weber. 2017.
\newblock Automated hate speech detection and the problem of offensive
  language.
\newblock In \emph{Proceedings of the International AAAI Conference on Web and
  Social Media}, volume~11.

\bibitem[{Facebook(2021)}]{fb_community_standards}
Facebook. 2021.
\newblock Community standards hate speech.
\newblock \url{https://www.facebook.com/communitystandards/hate_speech}.
\newblock Accessed on 12 June 2021.

\bibitem[{Founta et~al.(2018)Founta, Djouvas, Chatzakou, Leontiadis, Blackburn,
  Stringhini, Vakali, Sirivianos, and Kourtellis}]{founta2018large}
Antigoni Founta, Constantinos Djouvas, Despoina Chatzakou, Ilias Leontiadis,
  Jeremy Blackburn, Gianluca Stringhini, Athena Vakali, Michael Sirivianos, and
  Nicolas Kourtellis. 2018.
\newblock Large scale crowdsourcing and characterization of twitter abusive
  behavior.
\newblock In \emph{Proceedings of the International AAAI Conference on Web and
  Social Media}, volume~12.

\bibitem[{Gasparini et~al.(2018)Gasparini, Erba, Fersini, and
  Corchs}]{gasparini2018multimodal}
Francesca Gasparini, Ilaria Erba, Elisabetta Fersini, and Silvia Corchs. 2018.
\newblock Multimodal classification of sexist advertisements.
\newblock In \emph{ICETE (1)}, pages 565--572.

\bibitem[{{Jaded AI}()}]{easy}
{Jaded AI}.
\newblock Easy {OCR}.
\newblock \url{https://github.com/JaidedAI/EasyOCR}.

\bibitem[{Kiela et~al.(2020)Kiela, Firooz, Mohan, Goswami, Singh, Ringshia, and
  Testuggine}]{Kiela2020TheHM}
Douwe Kiela, Hamed Firooz, Aravind Mohan, Vedanuj Goswami, Amanpreet Singh,
  Pratik Ringshia, and Davide Testuggine. 2020.
\newblock The hateful memes challenge: Detecting hate speech in multimodal
  memes.
\newblock \emph{ArXiv}, abs/2005.04790.

\bibitem[{Kruk et~al.(2019)Kruk, Lubin, Sikka, Lin, Jurafsky, and
  Divakaran}]{Kruk2019IntegratingTA}
Julia Kruk, Jonah Lubin, Karan Sikka, X.~Lin, Dan Jurafsky, and Ajay Divakaran.
  2019.
\newblock Integrating text and image: Determining multimodal document intent in
  instagram posts.
\newblock \emph{ArXiv}, abs/1904.09073.

\bibitem[{MacAvaney et~al.(2019)MacAvaney, Yao, Yang, Russell, Goharian, and
  Frieder}]{macavaney2019hate}
Sean MacAvaney, Hao-Ren Yao, Eugene Yang, Katina Russell, Nazli Goharian, and
  Ophir Frieder. 2019.
\newblock Hate speech detection: Challenges and solutions.
\newblock \emph{PloS one}, 14(8):e0221152.

\bibitem[{Mathew et~al.(2018)Mathew, Kumar, Goyal, Mukherjee
  et~al.}]{mathew2018}
Binny Mathew, Navish Kumar, Pawan Goyal, Animesh Mukherjee, et~al. 2018.
\newblock Analyzing the hate and counter speech accounts on twitter.
\newblock \emph{arXiv preprint arXiv:1812.02712}.

\bibitem[{Nie et~al.(2021)Nie, Davani, Mathias, Kiela, Waseem, Vidgen, and
  Prabhakaran}]{nie_2021}
Shaoliang Nie, Aida Davani, Lambert Mathias, Douwe Kiela, Zeerak Waseem, Bertie
  Vidgen, and Vinodkumar Prabhakaran. 2021.
\newblock Woah shared task fine grained hateful memes classification.
\newblock
  \url{https://github.com/facebookresearch/fine_grained_hateful_memes/}.

\bibitem[{Pinterest(2021)}]{pinterest_about_this_page}
Pinterest. 2021.
\newblock All about pinterest.
\newblock \url{https://help.pinterest.com/en-gb/guide/all-about-pinterest}.
\newblock Accessed on 12 June 2021.

\bibitem[{Poria et~al.(2016)Poria, Cambria, Howard, Huang, and
  Hussain}]{poria2016fusing}
Soujanya Poria, Erik Cambria, Newton Howard, Guang-Bin Huang, and Amir Hussain.
  2016.
\newblock Fusing audio, visual and textual clues for sentiment analysis from
  multimodal content.
\newblock \emph{Neurocomputing}, 174:50--59.

\bibitem[{Radford et~al.(2021)Radford, Kim, Hallacy, Ramesh, Goh, Agarwal,
  Sastry, Askell, Mishkin, Clark, Krueger, and Sutskever}]{radford2021learning}
Alec Radford, Jong~Wook Kim, Chris Hallacy, Aditya Ramesh, Gabriel Goh,
  Sandhini Agarwal, Girish Sastry, Amanda Askell, Pamela Mishkin, Jack Clark,
  Gretchen Krueger, and Ilya Sutskever. 2021.
\newblock \href {http://arxiv.org/abs/2103.00020} {Learning transferable visual
  models from natural language supervision}.

\bibitem[{Rothe et~al.(2015)Rothe, Timofte, and Van~Gool}]{dex}
Rasmus Rothe, Radu Timofte, and Luc Van~Gool. 2015.
\newblock \href {https://doi.org/10.1109/ICCVW.2015.41} {Dex: Deep expectation
  of apparent age from a single image}.
\newblock In \emph{2015 IEEE International Conference on Computer Vision
  Workshop (ICCVW)}, pages 252--257.

\bibitem[{Sap et~al.(2019)Sap, Card, Gabriel, Choi, and Smith}]{sap2019risk}
Maarten Sap, Dallas Card, Saadia Gabriel, Yejin Choi, and Noah~A Smith. 2019.
\newblock The risk of racial bias in hate speech detection.
\newblock In \emph{Proceedings of the 57th annual meeting of the association
  for computational linguistics}, pages 1668--1678.

\bibitem[{Schroff et~al.(2015)Schroff, Kalenichenko, and Philbin}]{facenet}
Florian Schroff, Dmitry Kalenichenko, and James Philbin. 2015.
\newblock \href {https://doi.org/10.1109/CVPR.2015.7298682} {Facenet: A unified
  embedding for face recognition and clustering}.
\newblock In \emph{2015 IEEE Conference on Computer Vision and Pattern
  Recognition (CVPR)}, pages 815--823.

\bibitem[{Smith(2007)}]{smith2007overview}
Ray Smith. 2007.
\newblock An overview of the tesseract ocr engine.
\newblock In \emph{Ninth international conference on document analysis and
  recognition (ICDAR 2007)}, volume~2, pages 629--633. IEEE.

\bibitem[{Suryawanshi et~al.(2020)Suryawanshi, Chakravarthi, Arcan, and
  Buitelaar}]{suryawanshi2020multimodal}
Shardul Suryawanshi, Bharathi~Raja Chakravarthi, Mihael Arcan, and Paul
  Buitelaar. 2020.
\newblock Multimodal meme dataset (multioff) for identifying offensive content
  in image and text.
\newblock In \emph{Proceedings of the Second Workshop on Trolling, Aggression
  and Cyberbullying}, pages 32--41.

\bibitem[{Terhorst et~al.(2021)Terhorst, Kolf, Huber, Kirchbuchner, Damer,
  Morales, Fierrez, and Kuijper}]{Terhorst2021ACS}
P.~Terhorst, Jan~Niklas Kolf, Marco Huber, Florian Kirchbuchner, N.~Damer,
  A.~Morales, Julian Fierrez, and Arjan Kuijper. 2021.
\newblock A comprehensive study on face recognition biases beyond demographics.
\newblock \emph{ArXiv}, abs/2103.01592.

\bibitem[{{Van Der Maaten} and Hinton(2008)}]{VanDerMaaten2008}
Laurens {Van Der Maaten} and Geoffrey Hinton. 2008.
\newblock {Visualizing Data using t-SNE}.
\newblock Technical report.

\bibitem[{Waseem and Hovy(2016)}]{waseem2016}
Zeerak Waseem and Dirk Hovy. 2016.
\newblock Hateful symbols or hateful people? predictive features for hate
  speech detection on twitter.
\newblock In \emph{Proceedings of the NAACL student research workshop}, pages
  88--93.

\bibitem[{Zampieri et~al.(2019)Zampieri, Malmasi, Nakov, Rosenthal, Farra, and
  Kumar}]{zampieri2019}
Marcos Zampieri, Shervin Malmasi, Preslav Nakov, Sara Rosenthal, Noura Farra,
  and Ritesh Kumar. 2019.
\newblock Predicting the type and target of offensive posts in social media.
\newblock \emph{arXiv preprint arXiv:1902.09666}.

\bibitem[{Zhong(2020)}]{zhong2020classification}
Xiayu Zhong. 2020.
\newblock Classification of multimodal hate speech--the winning solution of
  hateful memes challenge.
\newblock \emph{arXiv preprint arXiv:2012.01002}.

\bibitem[{Zhou et~al.(2017)Zhou, Yao, Wen, Wang, Zhou, He, and Liang}]{east}
Xinyu Zhou, Cong Yao, He~Wen, Yuzhi Wang, Shuchang Zhou, Weiran He, and Jiajun
  Liang. 2017.
\newblock East: an efficient and accurate scene text detector.
\newblock In \emph{Proceedings of the IEEE conference on Computer Vision and
  Pattern Recognition}, pages 5551--5560.

\bibitem[{Zhou and Chen(2020)}]{zhou2020multimodal}
Yi~Zhou and Zhenhao Chen. 2020.
\newblock Multimodal learning for hateful memes detection.
\newblock \emph{arXiv preprint arXiv:2011.12870}.

\end{thebibliography}

\newpage
\onecolumn
\appendix

\section{Details on Pinterest Data Collection}
\label{sec:appendix_data}
Tab.~\ref{tab:pinterest_keywords_appendix} shows the keywords we use to search for memes on Pinterest. The search function returns images based on user-defined tags and descriptions aligning with the search term \cite{pinterest_about_this_page}. Each keyword search returns several hundred images on the first few pages of results. Note that Pinterest bans searches for `racist' memes or slurs associated with racial hatred so these could not be collected. We prefer this method of `noisy' labelling over classifying the memes with existing hate speech classifiers with the text as input because users likely take the multimodal content of the meme into account when adding tags or writing descriptions. However, we recognize that user-defined labelling comes with its own limitations of introducing noise into the dataset from idiosyncratic interpretation of tags. We also recognize that the memes we collect from Pinterest do not represent all Pinterest memes, nor do they represent all memes generally on the Internet. Rather, they reflect a sample of instances. Further, we over-sample non-hateful memes as compared to hateful memes because this distribution is one that is reflected in the real world. For example, the \fb~dev set is composed of 37\% hateful memes. Lastly, while we manually confirm that the noisy labels of 50 hateful and 50 non-hateful memes (see Tab.~\ref{tab:manual_check}), we also recognize that not all of the images accurately match the associated noisy label, especially for hateful memes which must match the definition of hate speech as directed towards a protected category.

\begin{table}[H]
\centering
\footnotesize
\caption{Keywords used to produce noisily-labelled samples of hateful and non-hateful memes from Pinterest.}
\label{tab:pinterest_keywords_appendix}
\begin{tabular}{ll} 
\toprule
\textbf{Noisy Label} & \textbf{Keywords}                                                                                                                                                 \\ 
\hline
Hate                 & ``sexist'', ``offensive'', ``vulgar'', ``wh*re'', ``sl*t'', ``prostitute''                                                     \\
Non-Hate             & ``funny'', ``wholesome'', ``happy'', ``friendship'', ``cute'', ``phd'', ``student'', ``food'', ``exercise''  \\
\bottomrule
\end{tabular}
\end{table}

\begin{table}[H]
\centering
\footnotesize
\caption{Results of manual annotation for noisy labelling. Of 50 random memes with a noisy hate label, we find 80\% are indeed hateful, and of 50 random memes with a noisy non-hate label, we find 94\% are indeed non-hateful.}
\label{tab:manual_check}
\begin{tabular}{lcc} 
\toprule
& \textbf{Noisy Hate} & \textbf{Noisy Non-Hate}    \\
\hline
\textbf{Annotator Hate} & 40 & 3 \\
\textbf{Annotator Non-Hate} & 10 & 47\\
\bottomrule
\end{tabular}
\end{table}

\section{Details on OCR Engines}
\label{sec:appendix_OCR}
\subsection{OCR Algorithms}
We evaluate three OCR algorithms on the \pin~and \fb~datasets. First, Tesseract \citep{smith2007overview} is Google’s open-source OCR engine. It has been continuously developed and maintained since its first release in 1985 by Hewlett-Packard Laboratories. Second, EasyOCR \citep{easy} developed by Jaded AI, is the algorithm used by the winner of the Facebook Hateful Meme Challenge. Third, East \citep{east} is an efficient deep learning algorithm for text detection in natural scenes. In this paper East is used to isolate regions of interest in the image in combination with Tesseract for text recognition.

\subsection{OCR Pre-filtering}
\begin{figure}[H]
    \centering
    \includegraphics[width=.6\textwidth]{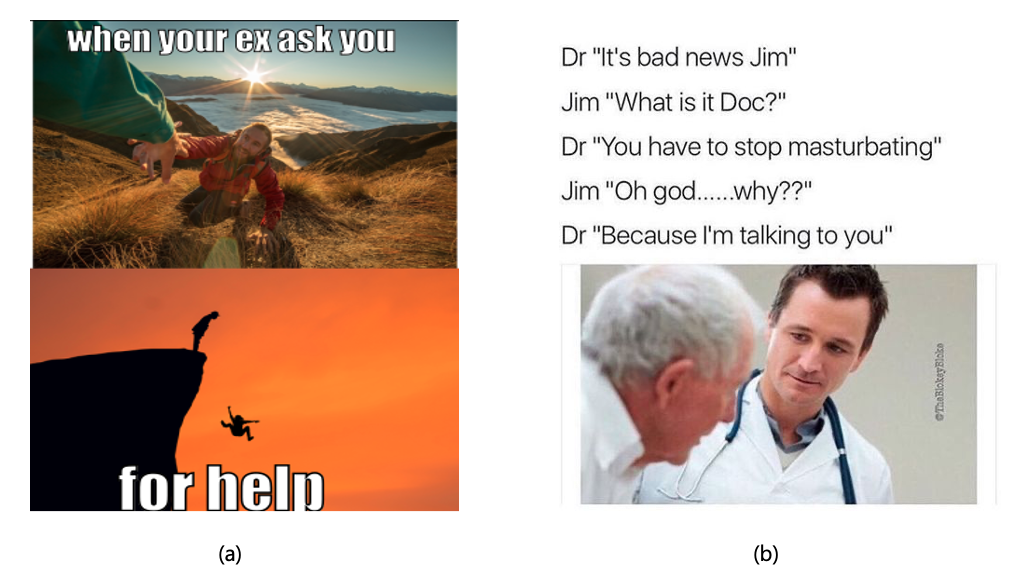}
    \caption{Dominant text patterns in (a) Facebook dataset (b) Pinterest dataset.}
    \label{fig:meme1}
\end{figure}

Figure \ref{fig:meme1} shows the dominant text patterns in \fb~(a) and \pin~(b) datasets, respectively. We use a specific prefiltering adapted to each pattern as follows. \\

\textbf{FB Tuning:} \fb~memes always have a black-edged white Impact font. The most efficient prefiltering sequence consists of applying an RGB-to-Gray conversion, followed by binary thresholding, closing, and inversion.
\textbf{Pin Tuning:} \pin~memes are less structured than \fb~memes, but a commonly observed meme type is text placed outside of the image on a uniform background. For this pattern, the most efficient prefiltering sequence consists of an RGB-to-Gray conversion followed by Otsu’s thresholding.\\

The optimal thresholds used to classify pixels in binary and Otsu's thresholding operations are found so as to maximise the average cosine similarity of the joint TF-IDF vectors between the labelled and predicted text from a sample of 30 annotated images from both datasets.

\section{Classification of Memes into Types}
\label{app:train_classifier}
\subsection{Data Preparation}
\begin{figure}[H]
    \centering
    \includegraphics[width=.8\textwidth]{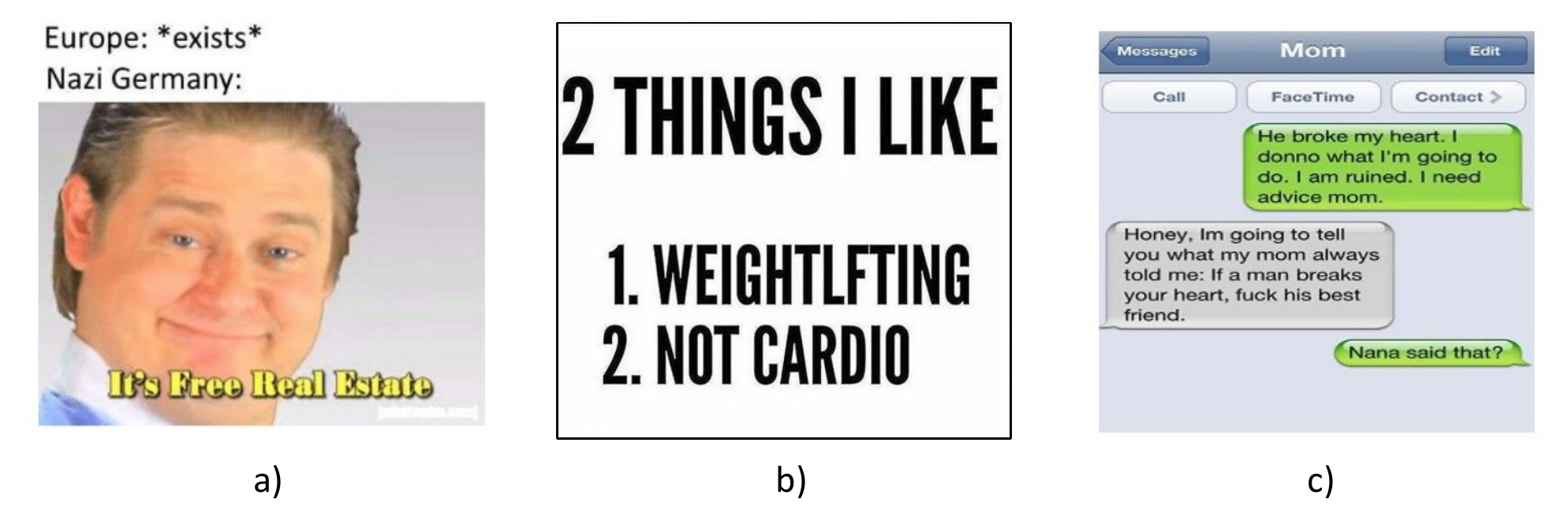}
    \caption{Different types of memes: (a) Traditional meme (b) Text (c) Screenshot.}
    \label{fig:meme2}
\end{figure}

To prepare the data needed for training the ternary (i.e., traditional memes, memes purely consisting of text, and screenshots) classifier, we annotate the \pin~dataset with manual annotations to create a balanced set of 400 images. We split the set randomly, so that 70\% is used as the training data and the rest 30\% as the validation data. Figure \ref{fig:meme2} shows the main types of memes encountered. The \fb~dataset only has traditional meme types. 

\subsection{Training Process}
We use image features taken from the penultimate layer of CLIP. We train a neural network with two hidden layers of 64 and 12 neurons respectively with ReLU activations, using Adam optimizer, for 50 epochs. The model achieves 93.3\% accuracy on the validation set.

\section{Classification Using CLIP}
\label{app:train_clip}
\subsection{Zero-shot Classification}
To perform zero-shot classification using CLIP \citep{radford2021learning}, for every meme we use two prompts, ``{\ttfamily a meme}" and ``{\ttfamily a hatespeech meme}". We measure the similarity score between the image and text embeddings and use the corresponding text prompt as a label. Note we regard this method as neither multimodal nor uni-modal, as the text is not explicitly given to the model, but as shown in \citep{radford2021learning}, CLIP has some OCR capabilities. In a future work we would like to explore how to modify the text prompts to improve performance.\\
\subsection{Linear Probing}
We train a binary linear classifier on the image features of CLIP on the \fb~train set. We train the classifier following the procedure outlined by \cite{radford2021learning}. Finally, we evaluate the binary classifier of the \fb~dev set and the \pin~dataset.\\

\noindent In all experiments above we use the pretrained ViT-B/32 model.

\section{Common Bigrams}
\label{app:bigrams}
The \fb~and \pin~datasets have distinctively different bigrams after data cleaning and the removal of stop words.

The most common bigrams for hateful \fb~memes are: [`black people', `white people', `white trash', `black guy', `sh*t brains', `look like']. The most common bigrams for non-hateful \fb~memes are: [`strip club', `isis strip', `meanwhile isis', `white people', `look like', `ilhan omar']

The most common bigrams for hateful \pin~memes are: `im saying', `favorite color', `single white', `black panthers', `saying wh*res', and `saying sl*t'. The most common bigrams for non-hateful \pin~memes are: `best friend', `dad jokes', `teacher new', `black lives', `lives matter', and `let dog'. 

\section{T-SNE Text Embeddings}
\label{app:tsne}
\begin{figure}[ht]
    \centering
    \includegraphics[width=0.4\columnwidth]{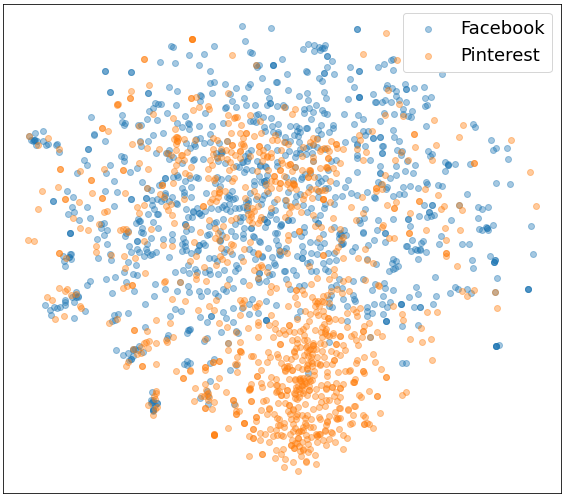} 
    \caption{t-SNE of Facebook and Pinterest memes' text-embeddings for a random sample of 1000 each.}
    \label{fig:tsne}
\end{figure}

The meme-level embeddings are calculated by (i) extracting a 300-dimensional embedding for each word in the meme, using fastText embeddings trained on Wikipedia and Common Crawl; (ii) averaging all the embeddings along each dimension. A T-SNE transformation is then applied to the full dataset, reducing it to two-dimensional space. After this reduction, 1000 text-embeddings from each category---\fb~and \pin~--- are extracted and visualized. The default perplexity parameter of 50 is used. 
Fig.\ref{fig:tsne} presents the t-SNE plot \citep{VanDerMaaten2008}, which indicates a concentration of multiple embeddings of the \pin~memes within a region at the bottom of the figure. These memes represent those that have nonsensical word tokens from OCR errors. 

\section{Face Recognition}
\label{app:face_recognition}
\subsection{Multi-Faces Detection Method}
To evaluate memes with multiple faces, we develop a self-adaptive algorithm to separate faces.
For each meme, we enumerate the position of a cutting line (either horizontal or vertical) with fixed granularity, and run facial detection models on both parts separately.
If both parts have a high probability of containing faces, we decide that each part has at least one face. Hence, we cut the meme along the line, and run this algorithm iteratively on both parts.
If no enumerated cutting line satisfies the condition above, then we decide there's only one face in the meme and terminate the algorithm.

\subsection{Additional Results on Facial Analysis}

\begin{table}[!htbp]
\caption{Predicted ratio of emotion categories on faces from different datasets from pre-trained DEX model.}
\label{tab:appendix_facial_analysis_emotion}
\centering
\footnotesize
\begin{tabular}{l cccc}
\toprule
& \multicolumn{2}{c}{\textbf{\fb}} & \multicolumn{2}{c}{\textbf{\pin}}  \\ 
\textit{categories} & Hate   & Non-Hate   & Hate  & Non-Hate \\ 
\midrule 
angry   & 10.6\% & 10.1\% & 9.0\% & 13.7\% \\ 
disgust   & 0.3\% & 0.2\% & 0.7\% & 0.6\% \\ 
fear   & 9.5\% & 10.2\% & 10.6\% & 13.0\% \\ 
happy   & 35.1\% & 36.3\% & 34.2\% & 30.1\% \\ 
neutral   & 23.1\% & 22.7\% & 23.4\% & 21.5\% \\ 
sad   & 18.8\% & 18.7\% & 20.4\% & 18.6\% \\ 
surprise   & 2.2\% & 1.7\% & 1.7\% & 1.8\% \\ 
\bottomrule
\end{tabular}
% }
\end{table}

\begin{table}[!htbp]
\caption{Predicted ratio of racial categories of faces from different datasets from pre-trained DEX model.}
\label{tab:appendix_facial_analysis_demographic}
\centering
\footnotesize
\begin{tabular}{l cccc}
\toprule
& \multicolumn{2}{c}{\textbf{\fb}} & \multicolumn{2}{c}{\textbf{\pin}}  \\ 
\textit{categories} & Hate   & Non-Hate   & Hate  & Non-Hate \\ 
\midrule 
asian          & 10.6\% & 10.8\%& 9.7\%& 13.9\%\\
black          & 15.0\% & 15.3\%& 6.5\%& 11.0\%\\
indian         & 5.9\% & 6.1\%& 3.2\%& 5.1\% \\
latino hispanic& 14.3\%& 14.5\% &10.2\% &11.7\% \\
middle eastern & 12.7\%& 11.2\%&9.5\% &10.1\% \\
white          & 41.5\% &42.1\% &60.9\% &48.1\% \\

\bottomrule
\end{tabular}
% }
\end{table}

\subsection{Examples of Faces in Memes}

\label{sec:faces_example}
% Figure \ref{fig:meme1}  \ref{fig:meme2} 
\begin{figure}[H]
    \centering
    \begin{subfigure}[b]{0.22\textwidth}
         \centering
         \includegraphics[width=\textwidth]{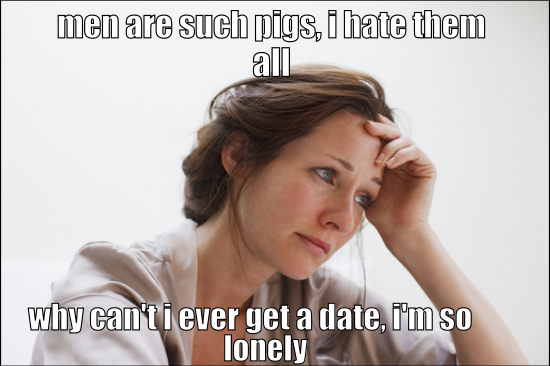}
         \caption{\fb~Hate}
         \label{fig:FBH}
    \end{subfigure}
    \begin{subfigure}[b]{0.22\textwidth}
         \centering
         \includegraphics[width=\textwidth]{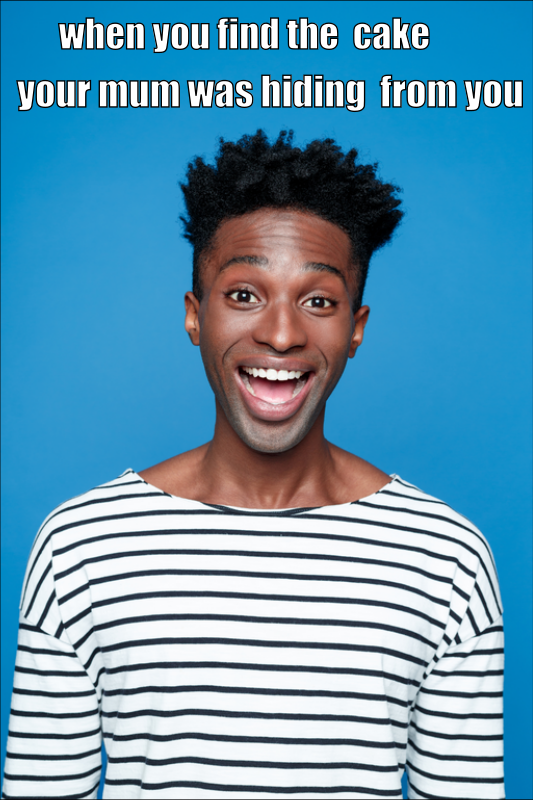}
         \caption{\fb~Non-hate}
         \label{fig:FBB}
    \end{subfigure}
    \begin{subfigure}[b]{0.22\textwidth}
         \centering
         \includegraphics[width=\textwidth]{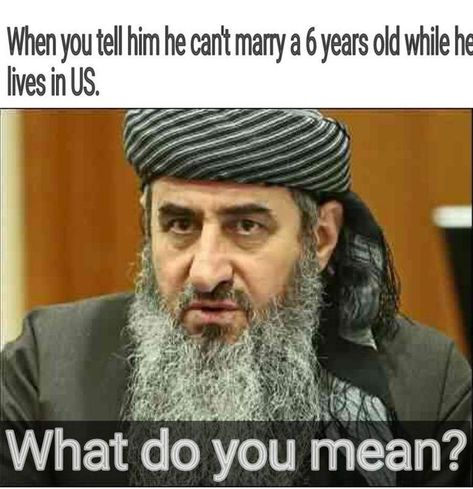}
         \caption{\pin~Hate}
         \label{fig:PH}
    \end{subfigure}
    \begin{subfigure}[b]{0.22\textwidth}
         \centering
         \includegraphics[width=\textwidth]{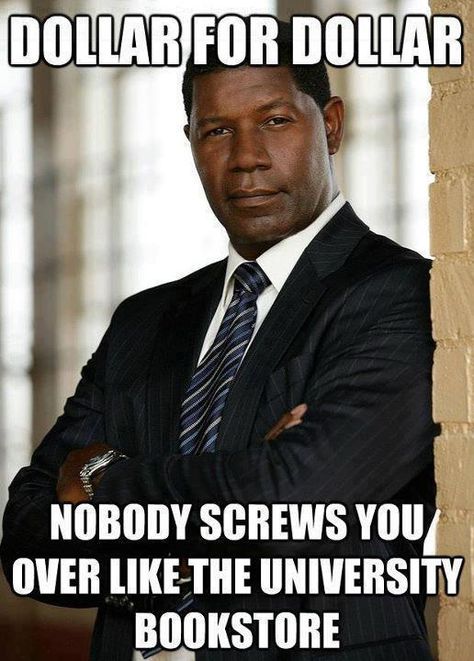}
         \caption{\pin~Non-hate}
         \label{fig:PB}
    \end{subfigure}
    \caption{Samples of faces in \fb~Hate, \fb~ Non-hate, \pin~Hate, and \pin~Non-hate datasets, and their demographic characteristic predicted by the DEX model:
    \newline (a) Woman, 37, white, sad (72.0\%); (b) Man, 27, black, happy (99.9\%); 
    \newline (c) Man, 36, middle eastern, angry (52.2\%); (d) Man, 29, black, neutral (68.0\%)}
    \label{fig:samples}
\end{figure}

\end{document}